# Dynamic Relation Inference via Verb Embeddings

Omri Suissa
Brown University
omri_suissa@brown.edu
omrishsu@gmail.com

Muhiim Ali
Brown University
muhiim_ali@brown.edu

Ariana Azarbal
Brown University
ariana_azarbal@brown.edu

Hui Shen
Brown University
hui_shen@alumni.brown.edu

Shekhar Pradhan
Brown University
shekhar_pradhan@brown.edu
shprad@gmail.com

## Abstract

*CLIP has demonstrated exceptional image-text matching capabilities due to its training on contrastive learning tasks. Past research has suggested that whereas CLIP effectively matches text to images when the matching can be achieved just by matching the text with the objects in the image, CLIP struggles when the matching depends on representing the relationship among the objects in the images (*i.e. *inferring relations). Previous attempts to address this limitation by training CLIP on relation detection datasets with only linguistic supervision have met with limited success. In this paper, we offer insights and practical methods to advance the field of relation inference from images. This paper approaches the task of creating a model that effectively detects relations among the objects in images by producing text and image embeddings that capture relationships through linguistic supervision. To this end, we propose Dynamic Relation Inference via Verb Embeddings (DRIVE), which augments the COCO dataset, fine-tunes CLIP with hard negatives subject-relation-object triples and corresponding images, and introduces a novel loss function to improve relation detection. Evaluated on multiple CLIP-based models, our method significantly improves zero-shot relation inference accuracy in both frozen and fine-tuned settings, significantly outperforming CLIP and state-of-the-art models while generalizing well on unseen data.*

## 1. Introduction

Vision-Language Pre-training (VLP) models have significantly advanced multimodal visual representation learning [34, 48, 57, 60, 63, 93, 111]. Among these, Contrastive Language-Image Pretraining (CLIP) [15, 83] has become a foundational model for further research [44, 60, 98, 111] demonstrating versatile cross-modal capabilities across tasks like object detection [36, 69, 106], image classification [83], text-to-image generation [99, 130], and semantic segmentation [7, 56, 110]. Despite its success, CLIP has limitations in capturing relational information within images [55, 115]. For example, when presented with an image and two captions that reference the same objects but differ in relationship or direction (*e.g.* "The horse is eating the grass" vs. "The grass is eating the horse"), CLIP often fails to select the correct caption. This suggests that CLIP primarily captures object presence rather than their relationships [115]. To address this challenge, we propose leveraging linguistic supervision to guide CLIP ([24]) in distinguishing relational differences among images containing the same objects but different relationships. Our goal is to train the model to recognize relations between objects in images (*e.g.* "a woman *pulling* a child" vs. "a woman *carrying* a child") and to understand differences in the directionality of relationships (*e.g.* "a *woman* pulling a *child*" vs. "a *child* pulling a *woman*"). We encourage the model to attend to relational information by training it on a simplified version of the captions, which emphasizes relational differences, and by introducing a novel loss function that accounts for relational composition differences (as detailed in Sec. 3). Our experiments (Sec. 4) demonstrate that our proposed *DRIVE* approach enhances CLIP's ability to infer relations, with our model *DriveCLIP* showing substantial improvements in relational understanding. Our key contributions are:

1. **Practical Strategy:** Proposing a novel method (*DRIVE*) for dataset generation that enhances relation inference through linguistic supervision (Sec. 3).
2. **New Benchmarks:** Introducing two COCO-based datasets (*CROCO* and *CROCO-D*) specifically designed for advancing relation inference tasks for the research community.

3. **New Loss Function:** Guiding the model to better infer relations by introducing a novel Hard Negative Loss function (Eq. (7)) that leverages linguistic supervision to improve contrastive learning (Sec. 3.2).
4. **Linguistic Insights:** Identifying specific limitations in CLIPs' ability to infer relationships depending on the linguistic representation of the relation (*i.e.* verb) state.

## 2. Related Work

Advancements in Vision-Language Pre-training (VLP) models [9, 34, 44, 46, 48, 60, 63, 93, 97, 107, 108, 111, 119, 123] have led to foundational models like Contrastive Language-Image Pretraining (CLIP) [15, 83], accelerating research in the field [37, 44, 60, 67, 71, 92, 98, 111, 122]. CLIP aligns images and textual annotations via contrastive learning [3, 12, 16, 74], but it struggles with visual relationship detection due to its lack of focus on object interactions [40, 55, 84, 100, 131], making it difficult to distinguish different relations among the same objects. In contrastive learning models, it is common to use hard negatives [23, 31, 88] (*i.e.* data points close to the anchor sample but belong to different classes) to learn nuanced distinctions between classes [47, 85]. Recently, hard negative mining using generative models has been shown to create high-quality hard negatives, improving VLMs' performance on compositional tasks [87]. In this context, our approach utilizes linguistically supervised hard negatives.

### 2.1. Data

The MS COCO dataset [5, 13, 65, 75, 78, 102] has significantly advanced computer vision research [6, 10, 11, 14, 20, 39, 62, 66, 82, 104, 114]. Like studies such as RefCOCO [113], we augment COCO with additional information, including images generated using DALL-E 3 [64]. Besides COCO, it is worth mentioning other datasets [45, 49, 70, 86, 90, 125] that focus on objects, attributes [1, 8, 27, 28, 52, 73, 91, 103, 105], and relations. Specifically, Visual Genome (VG) [50] provides extensive annotations such as objects, attributes, and relationships. Building on this, the Attribution, Relation, and Order (ARO) benchmark evaluates models' understanding of composition and order [115]. Building on prior works [80, 81] that decomposed texts into triplets (subject, predicate, object) using weak supervision, we use generative models to generate datasets for zero-shot relation inference. Whereas [117] combined visual and linguistic supervision with custom loss functions using visual cues like bounding boxes, we simplify text input using only linguistic supervision without manipulating the images to create hard negatives. Previous studies [22, 24, 41, 61, 77, 120, 124] introduced zero-shot visual classification methods, including relationship detection, using CLIP and custom loss functions, employing prompts to extend text with detailed descriptions. *In contrast, our approach simplifies textual descriptions to sharpen the model's attention, providing more precise linguistic supervision within this domain. Additionally, we enhance the contrastive loss function to better leverage linguistic supervision strategies.*

### 2.2. Relation Inference

Relations in natural language are often expressed as verbs linking subjects and objects in subject-verb-object (SVO) structures. Linguists distinguish between *stative* and *dynamic* verbs [53, 54]; stative verbs express a state (*e.g.* "a man *near* a building"), while dynamic verbs express actions or changes (*e.g.* "a man *enters* a building"). In this paper, we leverage this distinction in relations to capture the degree to which relations are represented in image embeddings, though captions in datasets like COCO and even datasets such as VL-CheckList [125] lack explicit differentiation between relations, instead presenting a blend of captions without clear categorization or with noisy (*i.e.* over-simplified) categorization (as detailed in Sec. 3.1). Relation inference (*i.e.* extraction) identifies structured facts from unstructured text using linguistic clues [38, 79, 127], often adopting a triplet format [18, 96, 128, 129], which we also use to represent sentences (*i.e.* captions).

Visual relationship inference (*i.e.* detection) [2, 19, 51, 118] aims to identify relationships between objects in images, recognizing subject-object pairs and their interactions. Previous works focus on spatial relationships [32, 35], human-object interaction [33, 126], and scene graph generation [42, 59, 109]. *We address the challenge of inferring visual relationships from real-world images across diverse relational types, and we provide insights into the linguistic features that influence model performance (Sec. 5.2).*

## 3. The DRIVE Method

To tackle the task of inferring relations from images using linguistic cues, we introduce both a new dataset and a novel loss function. Although existing datasets such as COCO [65] and its relational task variants like RefCOCO [113] are available, they do not offer the level of linguistic control required for our approach. To address this limitation, we generated fine-grained, relation-focused datasets that enable high-precision linguistic supervision. As shown in Fig. 1 and more explicitly in Algorithm 2, we employ our Dynamic Relation Inference via Verb Embeddings (*DRIVE*) method to generate the Contrastive Relations of Objects in COntext (*CROCO*) and its directional variant, Contrastive Relations of Objects in COntext with Directionality (*CROCO-D*), enabling the model to focus on relations during training (Sec. 3.1). Additionally, we propose *Drive-CLIP* (Sec. 3.2), a CLIP model trained with a new loss function that helps to improve relational inference (Sec. 3.2). As demonstrated in our experiments (Sec. 4), this approach significantly enhances CLIP's ability to infer relations.

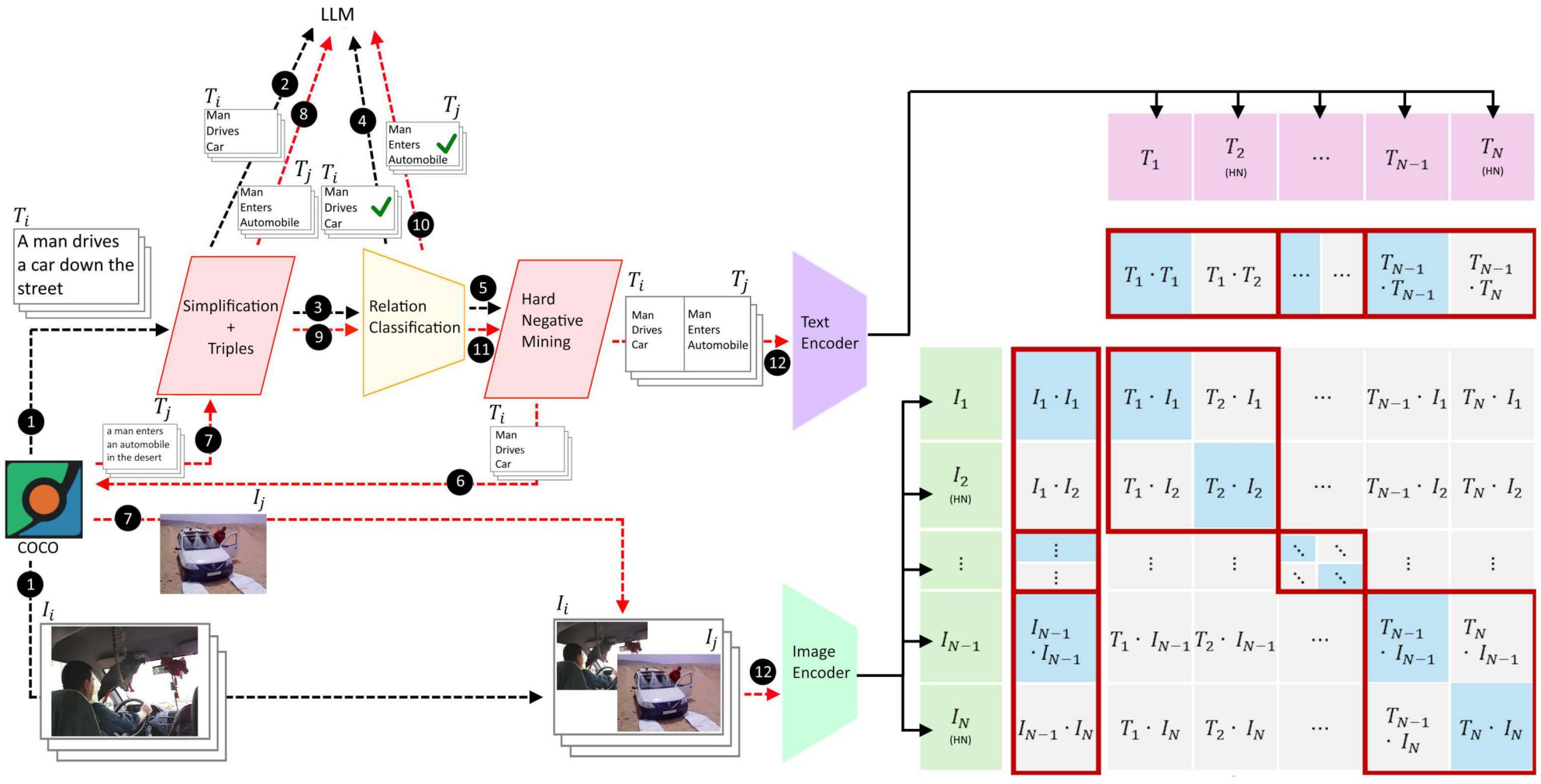

Figure 1. Overview of the DRIVE method. where $T_i$ and $I_i$ are image pairs from the COCO dataset, $T_j$ and $I_j$ are hard negative pair, $T_N$ is the text encoding of the $N$th caption, and $I_N$ is the image encoding of the corresponding image. The blue background denotes true labels, and gray indicates false labels. The red border highlights values used in the hard negative loss (Algorithm 2 and Eq. (7)).

## 3.1. Adapting COCO for Relation Inference

### 3.1.1. Linguistic Supervision

As we aim to precisely guide the image and text encoders - focusing solely on the relations between objects - we must have complete control over linguistic supervision. The first step in our *DRIVE* approach is to reshape the COCO dataset to achieve this level of control. We use the COCO dataset (also employed in the training of CLIP) and simplify the captions (*i.e.* texts) from sentences to subject-relation-object triples. This simplification allows us to control changes in the text and ensure that only the relation varies in the contrastive task.

**Algorithm 1** Caption Simplification and Triplet Extraction

**Require:** $T_n$

1: **if** $0 < Scene(T_n).rel < 3$ **and** $0 < T_n.obj \leq 3$ **then**
2: $\quad S \leftarrow LLM_{simplify}(T_n.caption)$ ▷ Fig. 1 steps (2,8)
3: $\quad s \leftarrow subject\text{-}head(S)$
4: $\quad r \leftarrow relation\text{-}span(S)$
5: $\quad o \leftarrow object\text{-}head(S)$
6: $\quad T_n.txt \leftarrow s + r + o$
7: $\quad T_n.state \leftarrow LLM_{state}(T_n.txt)$ ▷ Fig. 1 steps (4,10)
8: $\quad$ **return** $T_n$ ▷ Fig. 1 steps (5,11)
9: **end if**

where $T_n$ is an image-caption pair; $Scene(T).rel$ are relations from the Stanford Scene Graph Parser; $T_n.obj$ are COCO object annotations; $T_n.caption$ is the caption of $T_n$; $LLM_{simplify}$ splits a caption into a (subject, relation, object) triplet using LLM; $LLM_{state}$ classifies the verb as stative or dynamic using LLM; and *subject-head*, *relation-span*, *object-head* are SpaCy-based parsers.

As shown in Algorithm 1, we filter the COCO dataset by removing captions that lack relations or have more than two relations (identified using the Stanford Scene Graph Parser [89]) and images containing more than three main objects (based on COCO dataset annotations [65]). Next, we use an LLM (GPT-4o [43]) to simplify each caption to its main subject-relation-object triplet. We verify that the resulting caption has a valid subject-relation-object triple using SpaCy's part-of-speech tagger. We then concatenate the subject head, relation (*i.e.* verb) span, and direct object to generate the caption. [101]). We reduce captions to their fundamental subject-relation-object triplets so that we can more effectively mine hard negatives for linguistic supervision, which can lead the model to focus more closely on relation inference without being distracted by other factors. For example, when a model needs to choose between "A woman rides a skateboard in front of a large crowd" and "A woman in the park carries a skateboard" as a caption for an image. In this case, these captions provide multiple object cues (woman, skateboard, crowd, park), which can lead the model to predict based on the objects in the captions and lose focus on the difference in relations. However, simplifying the captions to "woman rides skateboard" versus "woman carries skateboard" forces the model to distinguish based solely on the relation since the subject and object are the same (*i.e.* woman $=$ woman, ride $\not\approx$ carry, skateboard $=$ skateboard). Hence the name Dynamic Relation Inference via *Verb Embeddings* as our linguistic control surrounds around verbs (*i.e.* relations).

Moreover, previous research demonstrates that positional and stative verbs (*i.e.* relations) pose challenges for inferring relations [55, 100]. We hypothesize that this difficulty arises because stative verbs depend heavily on perspective, allowing for multiple interpretations from the same image. For instance, the caption ”a man near a building” introduces ambiguity, as ”near” leaves room for different interpretations regarding the distance between the man and the building. In contrast, dynamic verbs are generally more explicitly represented, with less room for subjective interpretation. The visual information difference between two images with the same object and subject but with a different stative verb is generally low and comprises little change or a change that is based merely on rearranging patches of the same image. For example, the visual difference between the image expressed by ”his shirt resembles his pants” could be exactly the same as that expressed by the caption ”his shirt matches his pants”. Similarly, the visual difference between the image corresponding to ”a man to the left of a car” and the image corresponding to ”a man to the right of a car” consists simply of a change in the position of the two patches of the image, but not in any difference in what either object is doing. In contrast, the visual information difference between two images with the same object and subject but with a different dynamic verb is generally high and comprises of substantial change in the image. For example, the image of a man eating pasta will express substantially different visual information than that of a man cooking pasta. We believe that this assumption captures a significant intuition toward inferring relations using linguistic supervision that the image encoder used by CLIP (*e.g.* ResNet [121], ViT [21]) does not capture.

Moreover, we find that this distinction between stative and dynamic relations is essential and productive, and therefore, we aim to generate a dataset that will guide the model in this distinction. Based on this observation, the final step in the data processing part of our *DRIVE* approach splits the dataset into two subsets: $CROCO_s$, which contains stative relations, and $CROCO_d$, which contains dynamic relations. Since for some verbs whether they denote a stative or a dynamic relation depends on the entire context, the entire caption needs to be analyzed to determine the correct interpretation. Studies show that large language models (LLMs) can accurately classify verbs as stative or dynamic [17, 29, 30, 72, 76]. Building on this, we employ GPT-4o to classify each caption accordingly. In the COCO dataset, for example, the verb ”hold” was classified as dynamic in only $\approx 95\%$ of instances. A dynamic usage is seen in ”a man is holding a racquet,” while a stative usage appears in ”the clocktower holds lights”. This contextual approach offers better data than closed-list methods (*e.g.* [125] and is essential for applying linguistic supervision.

#### 3.1.2. Hard Negative Mining

As presented in Algorithm 2, using the data processing method of our *DRIVE* approach, we generated two datasets: Contrastive Relations of Objects in COntext (*CROCO*) and its directional variant, Contrastive Relations of Objects in COntext with Directionality (*CROCO-D*), both based on the MS Common Objects in COntext (MS COCO) dataset. These datasets are inspired by the notion of *Hard Negatives* [31], which have demonstrated the ability to improve contrastive language-image accuracy in past research [115]. However, the concept of hard negatives is broad, and we aim to provide more precise linguistic supervision. We define hard negative as a type of hard negative that focuses on one specific property of the text or image; in our case, the relation between objects. *CROCO* aims to produce a hard negative dataset to guide a CLIP model in distinguishing between differing semantic relations by changing only the relation within the caption. For example, as can be seen in Fig. 2a, the caption ”a man *driving* a car” is hard negative to ”a man *entering* a car” as the subject and the object are equal, and the relation is different. *CROCO-D* extends *CROCO* by adding the *directionality* of the relation, allowing the model to learn about plausibly asymmetric relationships. For example, as can be seen in Fig. 2b, the caption ”*Bride* feeds *groom*” is weakly directional contrastive to ”*Groom* feeds *bride*” as the object and the subject are reversed, and the relation is equal.

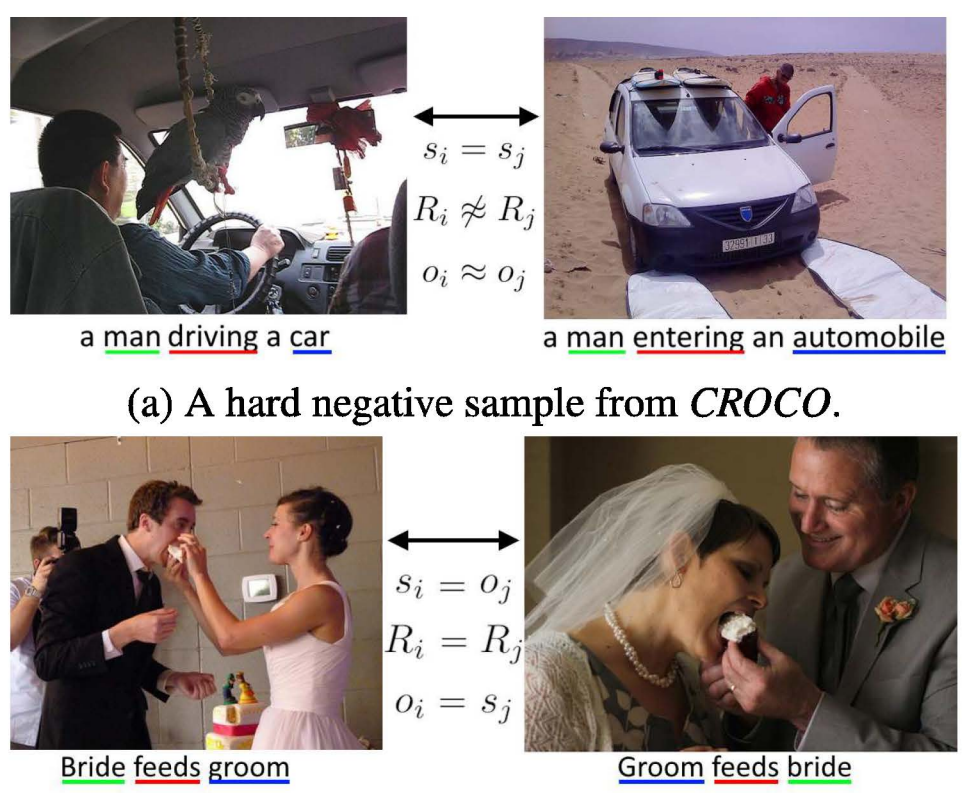

(a) A hard negative sample from *CROCO*.

(b) A hard negative sample from *CROCO-D*.

Figure 2. hard negative samples

Formally, we define a *triplet* (*i.e.* caption) as a 3-tuple consisting of a subject $(s)$, a relation $(R)$, and an object $(o)$. Each caption $(T)$ in our dataset can be represented as a triplet $(s, R, o)$. The *DRIVE* data processing phase ensures that all captions in the dataset can be decomposed into this simple form. As human language contains synonyms and semantically similar words, we define approximate equality for subjects $(s_1 \approx s_2)$ and objects $(o_1 \approx o_2)$ when these noun phrases are sufficiently synonymous based on SpaCy's synonyms classifier [101] with an $\epsilon$ threshold, or semantically similar based on cosine similarity with a threshold of $0.93$. The $0.93$ threshold was selected based on an F1 analysis of various thresholds in the $[0.8 \ldots 0.99]$ range, compared against a ground truth of 10,000 captions generated using an LLM (GPT4-o). Similarly, for relations: $R_1 \approx_{\mathrm{r}} R_2$ when these verb phrases are sufficiently semantically similar using the same threshold. For example, the relation of the caption "a man *driving* a car" is approximately equal to the relation of the caption "a man *steering* a car".

*CROCO* contains mappings from each caption ($T$) to the set of all captions that are hard negatives with it ($HN_T$). This hard negative set is defined as captions with subjects and objects synonymous with those in the anchor caption but with semantically differing relations. Formally, each image-caption pair $(I_i, T_i)$ in our dataset is mapped to its hard negative set ($HN_T$):

$$HN_T : (I_i, T_i : (s_i, R_i, o_i)) \rightarrow (I_j, T_j : (s_j, R_j, o_j)) \\ \text{where } s_i \approx s_j, R_i \not\approx R_j, o_i \approx o_j \tag{1}$$

We also compare triplets in their full form to ensure that hard negative captions describe semantically distinct relationships: $(s_i, R_i, s_o) \not\approx (s_j, R_j, o_j)$. This step is necessary because some relations may not be directly similar but convey similar meanings in specific contexts. For example, "plane *is in* the sky" is semantically similar to "plane *flies through* the sky," although "is in" and "flies through" are not synonymous. Two triplets are considered approximately equal ($\approx$) when the captions they represent, as well as their standardized forms, are approximately equal:

$$(I_i, T_i : (s_i, R_i, o_i)) \approx (I_j, T_j : (s_j, R_j, o_j)) \quad \text{if} \\ (s_i, R_i, o_i) \rightarrow (s, R_i, o), \\ (s_j, R_j, o_j) \rightarrow (s, R_j, o) \\ \text{where } s \in \{s_i, s_j\}, o \in \{o_i, o_j\} \tag{2}$$

Similarly, *CROCO-D* maps each image-caption pair to other hard negative image-caption pairs where the relation is synonymous, but the directionality is reversed:

$$HN_T : (I_i, T_i : (s_i, R_i, o_i)) \rightarrow (I_j, T_j : (s_j, R_j, o_j)) \\ \text{where } s_i \approx o_j, R_i \approx R_j, o_i \approx s_j \tag{3}$$

We restrict *CROCO-D* to image-caption pairs where the caption has a plausibly asymmetric relation. For example, the caption "woman lifts a man" is plausibly asymmetric as "man lifts a woman" is also plausible. In contrast, the caption "man eats pasta" is not plausibly asymmetric as pasta cannot eat a man (luckily for us). Because the COCO dataset lacked sufficient samples that satisfy the *CROCO-D* hard negative mining constraints, we enriched the *CROCO-D* dataset by selecting 1220 samples from COCO, which were classified as having animal or human subjects and objects, a bidirectional relationship, and an asymmetric relationship by an LLM (GPT-4o). For these samples, we synthetically generated reversed plausibly asymmetric captions (also using GPT-4o) and created corresponding images using DALL·E 3 [64]. To minimize the probability that generating images using DALL·E 3 will introduce potential bias into the dataset, we manually validated and filtered the generated images to samples that capture the complexity of real-world relationships.

By focusing on such relations, *CROCO-D* enables models to learn about the importance of directionality in relationships, enhancing their understanding of plausibly asymmetric actions and events.

This refined approach to dataset generation allows for more precise linguistic supervision in contrastive language-image training and, as shown in the Sec. 4, leads to models that better understand nuanced relationships in human language.

**Algorithm 2** DRIVE Dataset Generation

**Require:** $D$
1: **for** $S_i \in$ *COCO* **do** ▷ Fig. 1 step (1)
2: $T_i \leftarrow$ Algorithm 1($S_i$) ▷ Fig. 1 steps (2-5)
3: $HN_T \leftarrow [\ldots]$
4: **for** $S_j \in$ *COCO* **do** ▷ Fig. 1 steps (7)
5: $T_j \leftarrow$ Algorithm 1($S_j$) ▷ , Fig. 1 steps (8-11)
6: **if** $D =$ *CROCO* **then**
7: **if** Equation (2)($T_i, T_j$) $= False$ **then**
8: **if** Equation (1)($T_i, T_j$) $= True$ **then**
9: $HN_T$.append($T_j$)
10: **end if**
11: **end if**
12: **else if** $D =$ *CROCO-D* **then**
13: **if** Equation (3)($T_i, T_j$) $= True$ **then**
14: $HN_T$.append($T_j$)
15: **end if**
16: **end if**
17: **end for**
18: **yield** ($T_i, HN_T$) ▷ Fig. 1 step (12)
19: **end for**

where $D$ denotes the dataset type (*CROCO* or *CROCO-D*); $HN_T$ is the hard negative set of $T_i$.

## 3.2. The DriveCLIP Model

In the second component of our *DRIVE* approach for guiding contrastive language-image training for relation inference, we focus on improving the calculation of the model's loss function. We observed that the standard CLIP loss (Eq. (4)), which computes similarities between all captions and images within a batch, is suboptimal for our purposes because it does not account for the hard negative nature of

the training batch:

$$\mathcal{L}_{\text{CLIP}} = \frac{1}{2}\left(CE(s \cdot I \odot T^{\top}, y) + CE(s \cdot T \odot I^{\top}, y)\right) \tag{4}$$

where $I$ and $T$ are the image and text embeddings, $y$ is the class (row-column index), $CE$ denotes the cross-entropy loss ($-\sum_{i=1}^{N} y_i \log(p_i)$), and $s$ is a scale hyperparameter.

To address this issue, we introduce a novel loss function, *CROCO Contrastive Loss* ($\mathcal{L}_{\text{CROCO}}$), that better guides the model on hard negative tasks. To this end, we divide each batch into mini-batches, each containing a single caption-image pair and its corresponding hard negative set (*i.e.* image-caption pairs that are hard negative to the anchor caption). Unlike masking methods (*e.g.* [4]), our strategy enhances model efficiency by enabling the parallel computation of smaller contrastive matrices. Next, we calculate the CLIP contrastive loss only within each mini-batch, ignoring similarities across different mini-batches (see Fig. 1).

$$\mathcal{L}_{\text{CROCO}} = \mathcal{L}_{\text{CLIP}}([T_N, HN_{T_N}], [I_N, HN_{I_N}]) \tag{5}$$

where $\mathcal{L}_{\text{CLIP}}$ is the CLIP loss (Eq. (4)) computed over the embeddings of the anchor caption $T_N$ and its hard negative captions $HN_{T_N}$, and the anchor image $I_N$ and its hard negative images $HN_{I_N}$ within the mini-batch.

By ignoring the contrastive loss between non-contrastive samples, we focus the model on the linguistic supervision provided by the *CROCO* and *CROCO-D* datasets (Sec. 3.1). Unlike prior efforts (*i.e.* loss functions) that primarily focused on improving CLIP's ability to distinguish between image-text pairs by exploring various methods of pair differentiation (e.g., [41, 58, 116, 124]), our approach emphasizes the hard negative samples within themselves. Specifically, we introduce two separate similarity losses: one between the primary caption and its hard negative captions ($\mathcal{L}_{HN_{\text{Text}}}$), and another between the primary image and its hard negative images ($\mathcal{L}_{HN_{\text{Image}}}$).

$$\begin{aligned}
\mathcal{L}_{HN_{\text{Text}}} &= CE\left(S\left(\left\{\cos(T_N, HN_{T_N}^{(n)}) \mid n \in HN_{T_N}\right\}\right)\right) \\
\mathcal{L}_{HN_{\text{Image}}} &= CE\left(S\left(\left\{\cos(I_N, HN_{I_N}^{(n)}) \mid n \in HN_{I_N}\right\}\right)\right)
\end{aligned} \tag{6}$$

where $CE$ is the cross-entropy loss, cos denotes cosine similarity ($\cos = \frac{x \cdot y}{|x||y|}$), $S$ is the sigmoid function ($\frac{1}{1+e^{-x}}$), $T_N$ and $I_N$ represent the anchor caption and image, $HN_{T_N}$ and $HN_{I_N}$ are sets of hard negative captions and images, and $HN_{T_N}^{(n)}$ and $HN_{I_N}^{(n)}$ are their $n$-$th$ elements.

This approach teaches the model about the semantic differences between the captions themselves, independent of the images, and vice versa. Using the mini-batch's $\mathcal{L}_{\text{CROCO}}$, $\mathcal{L}_{HN_{\text{Text}}}$, and $\mathcal{L}_{HN_{\text{Image}}}$, we compute the *Hard Negative Loss* ($\mathcal{L}_{\text{HN}}$) with hyperparameters to control the weight (*i.e.* importance) of the hard negative similarities:

$$\mathcal{L}_{\text{HN}} = \frac{\mathcal{L}_{\text{CROCO}} \cdot \left(1 + \frac{1}{2}\left(\delta_T \mathcal{L}_{HN_{\text{Text}}} + \delta_I \mathcal{L}_{HN_{\text{Image}}}\right)\right)}{2} \tag{7}$$

where $\delta_T$ and $\delta_I$ are hyperparameters scaling text and image hard negative similarities, respectively.

Unlike previous approaches that integrate similarity loss into CLIP by simply adding it to the contrastive loss [22, 24, 120], we scale $\mathcal{L}_{\text{CROCO}}$ with a hard negative similarity to more effectively supports the hard negative objective. This insight is based on the idea that high $\mathcal{L}_{\text{CROCO}}$ increases the importance of the similarity differences between the anchor captions and their hard negative counterparts ($\mathcal{L}_{HN_{\text{Text}}}$), enhancing model learning (and vice versa). Furthermore, in contrast to other methods (*e.g.* [120]), we also consider the differences between the anchor images and their hard negative versions ($\mathcal{L}_{HN_{\text{Image}}}$), thereby further enhancing the impact on the image encoder. The total batch loss is the sum of all $\mathcal{L}_{\text{HN}}$ losses from the mini-batches. This refined loss function better captures relationships in the data, leading to improved performance on relation inference tasks.

## 4. Experimental Setting

As detailed in 3.1, we generated two datasets, *CROCO* and *CROCO-D* for evaluation.

| Measure | *CROCO$_{d/s}$* | *CROCO-D* |
|---|---|---|
| Captions | 47,492 / 42,477 | 2,362 |
| Distinct Relations | 1,118 / 332 | 170 |
| $\overline{R}$ Frequency | 42.48 / 127.94 | 13.89 |
| $\sigma_R$ Frequency | 251.35 / 885.64 | 38.96 |
| Distinct Entities | 4,172 / 4,854 | 428 |
| $\overline{R}$ per Entity | 5.33 / 3.11 | 4.16 |
| $\sigma_R$ per Entity | 16.70 / 5.36 | 8.69 |

Table 1. Dataset Statistics

where *CROCO$_{d/s}$* are the two subsets of *CROCO* (*i.e.* dynamic and stative)

As shown in Tab. 1, after filtering captions that contained either no relations/objects or too many relations/objects (as described in Algorithm 1) from the *COCO* dataset and selecting one caption per image, we narrowed it down to approximately 90,000 image-caption pairs, thereby forming the *CROCO* dataset. To create *CROCO-D*, we mined 2,560 plausible asymmetric relations from COCO and generated the corresponding hard negative set. We fine-tuned the models on the *CROCO* and *CROCO-D* datasets using OpenClip [15] ViT-B/32 (151M parameters). Each dataset was randomly split into 80% training and 20% testing subsets. During training, Bayesian hyperparameter search [95] was done to identify optimal training parameters (Tab. 2). Training sessions were done on a single GPU card (H100) and were completed in $\sim$60 minutes over seven epochs, highlighting the efficiency of *DRIVE* in enhancing relation inference with minimal resources.

| Hyper-Parameter | Search Range | Selected Value |
|---|---|---|
| Batch size | $[32 \ldots 256]$ | 64 |
| Learning rate | $[1e{-}04 \ldots 1e{-}06]$ | $1e{-}05$ |
| $\delta_I$ | $[0.1 \ldots 2.0]$ | 1.223 |
| $\delta_T$ | $[0.1 \ldots 2.0]$ | 0.615 |

Table 2. Hyper Parameters Search

# 5. Results

As shown in Sec. 4, *DriveCLIP* significantly outperforms *CLIP*, previous hard negatives training methods, and other CLIP-based state-of-the-art models evaluated in this paper, including models that are larger in size (*e.g.* DFN [25]). These results demonstrate that *DriveCLIP* improves accuracy (R@1) (*i.e.* predicting the current caption/image first) in both text-to-image and image-to-text classification tasks. However, since the *DRIVE* method relies on linguistic supervision (*i.e.* manipulating the captions), the accuracy gain is more pronounced when classifying images based on texts, as this supervision is expected to enhance the image encoder with new information more than the text encoder. This is further supported by the scaling hyperparameters of the *Hard Negative Loss* (Eq. (7)), where $\delta_I$ (1.2) was found to be twice as important as $\delta_T$ (0.6).

To ensure that the *DRIVE* method does not overfit our datasets and effectively enhances CLIP's accuracy on unseen data (*i.e.* generalization test), we evaluate the models using the ARO Benchmark [115] for relations, which is a subset of the Visual Genome dataset [50] (denoted as *VG Relations [ARO]*). *VG Relations [ARO]* was selected as it is the only dataset in the ARO benchmark that includes relational samples. As demonstrated in Sec. 4 and Tab. 4, our method generalizes well on this unseen dataset, improving accuracy in both text-to-image and image-to-text tasks even though the *DriveCLIP* was not trained on *VG Relations [ARO]* while some of the other models were trained directly on it (*e.g.* CE-CLIP). When assessing performance on the *CROCO-D* dataset, the results reveal that CLIP-based state-of-the-art models struggle to capture relational directions (*i.e.* "man is eating pasta" vs. "pasta is eating a man") in both text-to-image and image-to-text tasks. This finding underscores the importance of the *CROCO-D* dataset for future research aimed at incorporating such relational understanding into vision-language models.

## 5.1. Ablation Study of Hard Negative Loss

To isolate the accuracy gain specifically attributable to the Hard Negative Loss (Eq. (7)) and to confirm that *DriveCLIP*'s superior relational inference is not merely due to being trained on our datasets, we fine-tuned the original CLIP model, EV-02, and CE-CLIP, the current best SOTA models for this task (EV-02 the best on the generalization test on VG Relations [ARO] and CE-CLIP on relation inference test on CROCO), on the same datasets as *DriveCLIP* (*i.e.* *CROCO* and *CROCO-D*). Additionally, as expected, the gain in image-to-text accuracy is lower compared to text-to-image and underscores the greater contribution of linguistic supervision to the image encoder compared to the text encoder. The results (Tab. 4) show that *DriveCLIP* still outperforms fine-tuned CLIP, EV-02, and CE-CLIP in both text-to-image and image-to-text accuracy measurements. This result demonstrates that the Hard Negative Loss effectively leverages linguistic supervision to guide the model in distinguishing between hard negative samples (*i.e.* inferring relations). Furthermore, the *DRIVE* method improved the accuracy of fine-tuned CLIP on our datasets when evaluated on the VG Relations [ARO] dataset, demonstrating the value of our datasets independently of our loss function.

Fig. 3 presents a visualization of the attention (mean across all heads) for models trained with and without the Hard Negative Loss. As shown, while the CLIP loss emphasizes the man's shoulder and, to a lesser extent, the dog's head, the Hard Negative Loss shifts the focus toward the man's entire hand, emphasizing the contact point with the dog, and the rest of the dog's body, hinting at the likely trajectory of the hand. It also accentuates the man's head and face, suggesting that the petting motion carries an emotional component. This pattern is consistently observed across most of the attention visualizations we randomly sampled.

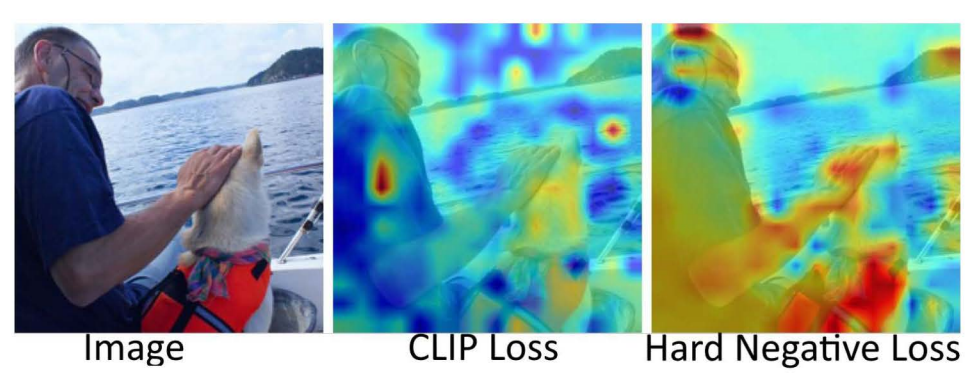

Figure 3. Visualization of the loss functions impact on the attention heads for the image-caption pair: "Man pets dog".

| | | HN | Text to Image | | | Image to Text | | |
|---|---|---|---|---|---|---|---|---|
| Model | Size | Training | *CROCO* ($aR_ib$ vs. $aR_jb$) | *CROCO-D* ($sRo$ vs. $oRs$) | *VG Relations [ARO]* (generalization test) | *CROCO* ($aR_ib$ vs. $aR_jb$) | *CROCO-D* ($sRo$ vs. $oRs$) | *VG Relations [ARO]* (generalization test) |
| *LaCLIP* [24] | 151M | X | 58.7% | 54.0% | 56.8% | 63.6% | 58.0% | 60.3% |
| *CoN-CLIP* [94] | 428M | ✓ | 58.7% | 50.8% | 57.4% | 66.7% | 58.4% | 65.7% |
| *CLIP* [83] | 151M | X | 59.4% | 52.3% | 57.6% | 64.9% | 54.4% | 64.2% |
| *DAC* [22] | 151M | ✓ | 59.5% | 50.2% | 57.0% | 67.7% | 53.1% | 64.4% |
| *COCA* [112] | 303M | X | 60.7% | 54.4% | 57.9% | 65.2% | 59.9% | 61.7% |
| *ConvNext* [68] | 198M | X | 62.2% | 53.1% | 59.2% | 66.8% | 57.8% | 64.1% |
| *EVA-02* [26] | 304M | X | 64.6% | 51.2% | 60.6% | 70.9% | 54.0% | 67.9% |
| *SigLIP* [116] | 878M | X | 64.8% | 53.1% | 60.1% | 66.1% | 59.4% | 64.2% |
| *NegCLIP* [115] | 151M | ✓ | 65.3% | 52.1% | 58.2% | 70.7% | 56.5% | 66.2% |
| *DFN* [25] | 632M | X | 65.9% | 55.6% | 60.5% | 69.7% | 59.4% | 65.8% |
| *CE-CLIP* [120] | 151M | ✓ | 68.7% | 55.6% | 58.9% | 73.6% | 55.4% | 66.8% |
| *DriveCLIP* [**ours**] | 151M | ✓ | **79.3%** | **81.5%** | **64.3%** | **81.0%** | **91.5%** | **69.5%** |

Table 3. Accuracy (R@1) on CROCO, CROCO-D, and Visual Genome Relations [ARO].

HN training indicates if the model was trained with hard negatives. *DriveCLIP* was not trained on VG Relations [ARO] and still outperforms SOTA models that were trained on VG Relations [ARO].

| | Text to Image | | | Image to Text | | |
|---|---|---|---|---|---|---|
| Model | *CROCO* ($aR_ib$ vs. $aR_jb$) | *CROCO-D* ($sRo$ vs. $oRs$) | *VG Relations [ARO]* (generalization test) | *CROCO* ($aR_ib$ vs. $aR_jb$) | *CROCO-D* ($sRo$ vs. $oRs$) | *VG Relations [ARO]* (generalization test) |
| *CLIP*$_{\textit{fine-tuned}}$ | 70.9% | 63.0% | 60.7% | 79.3% | 81.5% | 67.7% |
| *EVA-02*$_{\textit{fine-tuned}}$ | 72.6% | 73.8% | 60.3% | 80.1% | 81.0% | 67.0% |
| *CE-CLIP*$_{\textit{fine-tuned}}$ | 70.8% | 72.3% | 58.3% | 77.8% | 79.9% | 65.9% |
| *DriveCLIP* | **79.3%** | **81.5%** | **64.3%** | **81.0%** | **91.5%** | **69.5%** |

Table 4. Accuracy (R@1) gain of *DriveCLIP* comparing to best SOTA models fine-tuned on CROCO and CROCO-D.

### 5.2. Analysis of Dynamic vs. Stative Verbs

| | Text to Image | | Image to Text | |
|---|---|---|---|---|
| Model | *CROCO*$_s$ | $\Delta acc$ | *CROCO*$_s$ | $\Delta acc$ |
| *CLIP*$_{\textit{fine-tuned}}$ | 58.3% | -21.6% | 63.5% | -24.8% |
| *EV-02*$_{\textit{fine-tuned}}$ | 58.1% | -24.9% | 62.4% | -28.3% |
| *CE-CLIP*$_{\textit{fine-tuned}}$ | 57.8% | -22.4% | 62.4% | -24.6% |
| *DriveCLIP* | **62.4%** | -27.0% | **64.5%** | -25.5% |

Table 5. Accuracy (R@1) on the stative subset (*CROCO*$_s$). $\Delta acc$ is the impact of the stative relation on the accuracy.

Furthermore, to test our hypothesis that the limited success in previous research on relation inference arises from not distinguishing between stative and dynamic relations, we trained the best SOTA models and *DriveCLIP* separately on the stative subset of *CROCO* and evaluated them on that subset. As shown in Tab. 5, our hypothesis was confirmed: both the absolute text-to-image accuracy and the relative gain from the Hard Negative Loss in the text-to-image task are lower when training and evaluating exclusively on stative relations compared to dynamic relations. These results demonstrate that the underlying joint (*i.e.* image and text) distribution of stative relations differs from that of dynamic relations and its impact on accuracy ($\Delta acc$). Moreover, EV-02 and CE-CLIP's lower accuracy suggests current datasets (*e.g.* VG Relations [ARO]) lack our distinction between stative and dynamic relations and do misguide models.

## 6. Conclusion and Discussion

This paper introduces Dynamic Relation Inference via Verb Embeddings (*DRIVE*), a novel method that leverages linguistic supervision to enhance relation inference from images (Sec. 3). As demonstrated in Sec. 5, the *DRIVE* significantly improves relation detection in both image-to-text and text-to-image accuracy compared to CLIP and other CLIP-based models and generalizes well on unseen benchmarks (VG Relations [ARO]). Our research underscores the importance and potential of incorporating linguistic supervision to enhance joint embedding models like CLIP. We highlight the necessity of fine-grained linguistic control in the dataset by manipulating the inference goal (*i.e.*, relations) while keeping other elements constant (*i.e.*, subject and object). Moreover, we show that tailored loss functions in the joint embedding space can improve accuracy by compelling the model to better distinguish the inference goal and reduce noise from irrelevant contrastive embeddings (*i.e.* non-hard negative). Furthermore, unlike resource-intensive methods, our *DRIVE* method is relatively lightweight, needing only about an hour of fine-tuning on a single GPU.

While significant work remains, advancing and refining relation inference from images holds the potential to enhance models with more human-like perception. A paper like this can shape future research and applications across various domains. For example, in robotics, understanding both objects and their interactions can boost decision-making and adaptability. In healthcare, action-based image recognition can improve care by anticipating needs and preventing accidents. However, it is also essential to recognize the *limitations* of the *DRIVE* method due to its reliance on linguistic supervision, making it prone to textual and visual ambiguities and highly dependent on context. Moreover, captions as high-level abstractions might not capture subtle or implicit relationships, limiting the model's performance.

Looking ahead, we aim to address these challenges through several planned research directions, including extending the *DRIVE* method to incorporate other single-property hard negatives, exploring hard negative loss functions to enhance the model's ability to jointly identify objects, relations, and attributes, and exploring methods to add hard negative adapters to enhance the composability of existing models. By pursuing these advancements, we hope to push the boundaries of relation inference and ensure its positive impact across various fields.

## 7. Acknowledgement

This research work was supported by research grants from the Data Science Institute of Brown University to Shekhar Pradhan and to UTRA awards from Brown University to Muhiim Ali and Ariana Azarbal.

## References

[1] Zeynep Akata, Florent Perronnin, Zaid Harchaoui, and Cordelia Schmid. Label-embedding for attribute-based classification. In *Proceedings of the IEEE conference on computer vision and pattern recognition*, pages 819–826, 2013. 2

[2] Haider Al-Tahan, Quentin Garrido, Randall Balestriero, Diane Bouchacourt, Caner Hazirbas, and Mark Ibrahim. Scaling vision-language models does not improve relational understanding: The right learning objective helps. In *Proceedings of the IEEE/CVF Conference on Computer Vision and Pattern Recognition*, pages 8353–8358, 2024. 2

[3] Jiawang Bai, Kuofeng Gao, Shaobo Min, Shu-Tao Xia, Zhifeng Li, and Wei Liu. Badclip: Trigger-aware prompt learning for backdoor attacks on clip. In *Proceedings of the IEEE/CVF Conference on Computer Vision and Pattern Recognition*, pages 24239–24250, 2024. 2

[4] Timo Bertram, Johannes Fürnkranz, and Martin Müller. Contrastive learning of preferences with a contextual infonce loss. *arXiv preprint arXiv:2407.05898*, 2024. 6

[5] Holger Caesar, Jasper Uijlings, and Vittorio Ferrari. Coco-stuff: Thing and stuff classes in context. In *Proceedings of the IEEE conference on computer vision and pattern recognition*, pages 1209–1218, 2018. 2

[6] Nicolas Carion, Francisco Massa, Gabriel Synnaeve, Nicolas Usunier, Alexander Kirillov, and Sergey Zagoruyko. End-to-end object detection with transformers. In *European conference on computer vision*, pages 213–229. Springer, 2020. 2

[7] Junbum Cha, Jonghwan Mun, and Byungseok Roh. Learning to generate text-grounded mask for open-world semantic segmentation from only image-text pairs. In *Proceedings of the IEEE/CVF Conference on Computer Vision and Pattern Recognition*, pages 11165–11174, 2023. 1

[8] Yu-Wei Chao, Yunfan Liu, Xieyang Liu, Huayi Zeng, and Jia Deng. Learning to detect human-object interactions. In *2018 ieee winter conference on applications of computer vision (wacv)*, pages 381–389. IEEE, 2018. 2

[9] Lianggangxu Chen, Xuejiao Wang, Jiale Lu, Shaohui Lin, Changbo Wang, and Gaoqi He. Clip-driven open-vocabulary 3d scene graph generation via cross-modality contrastive learning. In *Proceedings of the IEEE/CVF Conference on Computer Vision and Pattern Recognition*, pages 27863–27873, 2024. 2

[10] Liang-Chieh Chen, George Papandreou, Iasonas Kokkinos, Kevin Murphy, and Alan L Yuille. Deeplab: Semantic image segmentation with deep convolutional nets, atrous convolution, and fully connected crfs. *IEEE transactions on pattern analysis and machine intelligence*, 40(4):834–848, 2017. 2

[11] Liang-Chieh Chen, Yukun Zhu, George Papandreou, Florian Schroff, and Hartwig Adam. Encoder-decoder with atrous separable convolution for semantic image segmentation. In *Proceedings of the European conference on computer vision (ECCV)*, pages 801–818, 2018. 2

[12] Ting Chen, Simon Kornblith, Mohammad Norouzi, and Geoffrey Hinton. A simple framework for contrastive learning of visual representations. In *International conference on machine learning*, pages 1597–1607. PMLR, 2020. 2

[13] Xinlei Chen, Hao Fang, Tsung-Yi Lin, Ramakrishna Vedantam, Saurabh Gupta, Piotr Dollár, and C Lawrence Zitnick. Microsoft coco captions: Data collection and evaluation server. *arXiv preprint arXiv:1504.00325*, 2015. 2

[14] Bowen Cheng, Maxwell D Collins, Yukun Zhu, Ting Liu, Thomas S Huang, Hartwig Adam, and Liang-Chieh Chen. Panoptic-deeplab: A simple, strong, and fast baseline for bottom-up panoptic segmentation. In *Proceedings of the IEEE/CVF conference on computer vision and pattern recognition*, pages 12475–12485, 2020. 2

[15] Mehdi Cherti, Romain Beaumont, Ross Wightman, Mitchell Wortsman, Gabriel Ilharco, Cade Gordon, Christoph Schuhmann, Ludwig Schmidt, and Jenia Jitsev. Reproducible scaling laws for contrastive language-image learning. In *Proceedings of the IEEE/CVF Conference on Computer Vision and Pattern Recognition*, pages 2818–2829, 2023. 1, 2, 6

[16] Sumit Chopra, Raia Hadsell, and Yann LeCun. Learning a similarity metric discriminatively, with application to face verification. In *2005 IEEE Computer Society Conference*

*on Computer Vision and Pattern Recognition, CVPR 2005*, pages 539–546. IEEE Computer Society, 2005. 2

[17] Bernard Comrie. *Aspect: An introduction to the study of verbal aspect and related problems*. Cambridge University Press, 1976. 4

[18] Aron Culotta and Jeffrey Sorensen. Dependency tree kernels for relation extraction. In *Proceedings of the 42nd annual meeting of the association for computational linguistics (ACL-04)*, pages 423–429, 2004. 2

[19] Bo Dai, Yuqi Zhang, and Dahua Lin. Detecting visual relationships with deep relational networks. In *Proceedings of the IEEE conference on computer vision and Pattern recognition*, pages 3076–3086, 2017. 2

[20] Xueqing Deng, Qihang Yu, Peng Wang, Xiaohui Shen, and Liang-Chieh Chen. Coconut: Modernizing coco segmentation. In *Proceedings of the IEEE/CVF Conference on Computer Vision and Pattern Recognition*, pages 21863–21873, 2024. 2

[21] Alexey Dosovitskiy, Lucas Beyer, Alexander Kolesnikov, Dirk Weissenborn, Xiaohua Zhai, Thomas Unterthiner, Mostafa Dehghani, Matthias Minderer, Georg Heigold, Sylvain Gelly, et al. An image is worth 16x16 words: Transformers for image recognition at scale. In *International Conference on Learning Representations*, 2020. 4

[22] Sivan Doveh, Assaf Arbelle, Sivan Harary, Roei Herzig, Donghyun Kim, Paola Cascante-Bonilla, Amit Alfassy, Rameswar Panda, Raja Giryes, Rogerio Feris, et al. Dense and aligned captions (dac) promote compositional reasoning in vl models. *Advances in Neural Information Processing Systems*, 36:76137–76150, 2023. 2, 6, 8

[23] Fartash Faghri, David J Fleet, Jamie Ryan Kiros, and Sanja Fidler. Vse++: Improving visual-semantic embeddings with hard negatives. *arXiv preprint arXiv:1707.05612*, 2017. 2

[24] Lijie Fan, Dilip Krishnan, Phillip Isola, Dina Katabi, and Yonglong Tian. Improving clip training with language rewrites. *Advances in Neural Information Processing Systems*, 36, 2024. 1, 2, 6, 8

[25] Alex Fang, Albin Madappally Jose, Amit Jain, Ludwig Schmidt, Alexander T Toshev, and Vaishaal Shankar. Data filtering networks. In *NeurIPS 2023 Workshop on Distribution Shifts: New Frontiers with Foundation Models*. 7, 8

[26] Yuxin Fang, Quan Sun, Xinggang Wang, Tiejun Huang, Xinlong Wang, and Yue Cao. Eva-02: A visual representation for neon genesis. *Image and Vision Computing*, 149: 105171, 2024. 8

[27] Ali Farhadi, Ian Endres, Derek Hoiem, and David Forsyth. Describing objects by their attributes. In *2009 IEEE conference on computer vision and pattern recognition*, pages 1778–1785. IEEE, 2009. 2

[28] Vittorio Ferrari and Andrew Zisserman. Learning visual attributes. *Advances in neural information processing systems*, 20, 2007. 2

[29] Annemarie Friedrich and Alexis Palmer. Automatic prediction of aspectual class of verbs in context. In *Annual Meeting of the Association for Computational Linguistics*, 2014. 4

[30] Annemarie Friedrich, Nianwen Xue, and Alexis Palmer. A kind introduction to lexical and grammatical aspect, with a survey of computational approaches. In *Proceedings of the 17th Conference of the European Chapter of the Association for Computational Linguistics*, pages 599–622, Dubrovnik, Croatia, 2023. Association for Computational Linguistics. 4

[31] Damianos Galanopoulos and Vasileios Mezaris. Hard-negatives or non-negatives? a hard-negative selection strategy for cross-modal retrieval using the improved marginal ranking loss. In *Proceedings of the IEEE/CVF International Conference on Computer Vision*, pages 2312–2316, 2021. 2, 4

[32] Carolina Galleguillos, Andrew Rabinovich, and Serge Belongie. Object categorization using co-occurrence, location and appearance. In *2008 IEEE Conference on Computer Vision and Pattern Recognition*, pages 1–8. IEEE, 2008. 2

[33] Chen Gao, Jiarui Xu, Yuliang Zou, and Jia-Bin Huang. Drg: Dual relation graph for human-object interaction detection. In *Computer Vision–ECCV 2020: 16th European Conference, Glasgow, UK, August 23–28, 2020, Proceedings, Part XII 16*, pages 696–712. Springer, 2020. 2

[34] Kuofeng Gao, Yang Bai, Jindong Gu, Shu-Tao Xia, Philip Torr, Zhifeng Li, and Wei Liu. Inducing high energy-latency of large vision-language models with verbose images. In *The Twelfth International Conference on Learning Representations*. 1, 2

[35] Stephen Gould, Jim Rodgers, David Cohen, Gal Elidan, and Daphne Koller. Multi-class segmentation with relative location prior. *International journal of computer vision*, 80: 300–316, 2008. 2

[36] Xiuye Gu, Tsung-Yi Lin, Weicheng Kuo, and Yin Cui. Open-vocabulary object detection via vision and language knowledge distillation. In *International Conference on Learning Representations*, 2021. 1

[37] Andrey Guzhov, Federico Raue, Jörn Hees, and Andreas Dengel. Audioclip: Extending clip to image, text and audio. In *ICASSP 2022-2022 IEEE International Conference on Acoustics, Speech and Signal Processing (ICASSP)*, pages 976–980. IEEE, 2022. 2

[38] Xu Han, Tianyu Gao, Yankai Lin, Hao Peng, Yaoliang Yang, Chaojun Xiao, Zhiyuan Liu, Peng Li, Jie Zhou, and Maosong Sun. More data, more relations, more context and more openness: A review and outlook for relation extraction. In *Proceedings of the 1st Conference of the Asia-Pacific Chapter of the Association for Computational Linguistics and the 10th International Joint Conference on Natural Language Processing*, pages 745–758, 2020. 2

[39] Xiaotian Han, Yiqi Wang, Bohan Zhai, Quanzeng You, and Hongxia Yang. Coco is" all"you need for visual instruction fine-tuning. *arXiv preprint arXiv:2401.08968*, 2024. 2

[40] Nils Hoehing, Ellen Rushe, and Anthony Ventresque. What's left can't be right–the remaining positional incompetence of contrastive vision-language models. *arXiv preprint arXiv:2311.11477*, 2023. 2

[41] Xuming Hu, Junzhe Chen, Aiwei Liu, Shiao Meng, Lijie Wen, and Philip S Yu. Prompt me up: Unleashing the power

of alignments for multimodal entity and relation extraction. In *Proceedings of the 31st ACM International Conference on Multimedia*, pages 5185–5194, 2023. 2, 6

[42] Yufeng Huang, Jiji Tang, Zhuo Chen, Rongsheng Zhang, Xinfeng Zhang, Weijie Chen, Zeng Zhao, Zhou Zhao, Tangjie Lv, Zhipeng Hu, et al. Structure-clip: Towards scene graph knowledge to enhance multi-modal structured representations. In *Proceedings of the AAAI Conference on Artificial Intelligence*, pages 2417–2425, 2024. 2

[43] Raisa Islam and Owana Marzia Moushi. Gpt-4o: The cutting-edge advancement in multimodal llm. *Authorea Preprints*, 2024. 3

[44] Chao Jia, Yinfei Yang, Ye Xia, Yi-Ting Chen, Zarana Parekh, Hieu Pham, Quoc Le, Yun-Hsuan Sung, Zhen Li, and Tom Duerig. Scaling up visual and vision-language representation learning with noisy text supervision. In *International conference on machine learning*, pages 4904–4916. PMLR, 2021. 1, 2

[45] Justin Johnson, Ranjay Krishna, Michael Stark, Li-Jia Li, David Shamma, Michael Bernstein, and Li Fei-Fei. Image retrieval using scene graphs. In *Proceedings of the IEEE conference on computer vision and pattern recognition*, pages 3668–3678, 2015. 2

[46] Armand Joulin, Laurens Van Der Maaten, Allan Jabri, and Nicolas Vasilache. Learning visual features from large weakly supervised data. In *Computer Vision–ECCV 2016: 14th European Conference, Amsterdam, The Netherlands, October 11–14, 2016, Proceedings, Part VII 14*, pages 67–84. Springer, 2016. 2

[47] Yannis Kalantidis, Mert Bulent Sariyildiz, Noe Pion, Philippe Weinzaepfel, and Diane Larlus. Hard negative mixing for contrastive learning. *Advances in neural information processing systems*, 33:21798–21809, 2020. 2

[48] Wonjae Kim, Bokyung Son, and Ildoo Kim. Vilt: Vision-and-language transformer without convolution or region supervision. In *International conference on machine learning*, pages 5583–5594. PMLR, 2021. 1, 2

[49] Alexander Kirillov, Eric Mintun, Nikhila Ravi, Hanzi Mao, Chloe Rolland, Laura Gustafson, Tete Xiao, Spencer Whitehead, Alexander C Berg, Wan-Yen Lo, et al. Segment anything. In *Proceedings of the IEEE/CVF International Conference on Computer Vision*, pages 4015–4026, 2023. 2

[50] Ranjay Krishna, Yuke Zhu, Oliver Groth, Justin Johnson, Kenji Hata, Joshua Kravitz, Stephanie Chen, Yannis Kalantidis, Li-Jia Li, David A Shamma, et al. Visual genome: Connecting language and vision using crowdsourced dense image annotations. *International journal of computer vision*, 123:32–73, 2017. 2, 7

[51] Alina Kuznetsova, Hassan Rom, Neil Alldrin, Jasper Uijlings, Ivan Krasin, Jordi Pont-Tuset, Shahab Kamali, Stefan Popov, Matteo Malloci, Alexander Kolesnikov, et al. The open images dataset v4: Unified image classification, object detection, and visual relationship detection at scale. *International journal of computer vision*, 128(7): 1956–1981, 2020. 2

[52] Christoph H Lampert, Hannes Nickisch, and Stefan Harmeling. Learning to detect unseen object classes by between-class attribute transfer. In *2009 IEEE conference on computer vision and pattern recognition*, pages 951–958. IEEE, 2009. 2

[53] Beth Levin. Dative verbs a crosslinguistic perspective. *Lingvisticæ investigationes*, 31(2):285–312, 2008. 2

[54] Beth C Levin. Instrumental with and the control relation in english. 1979. 2

[55] Martha Lewis, Nihal Nayak, Peilin Yu, Jack Merullo, Qinan Yu, Stephen Bach, and Ellie Pavlick. Does clip bind concepts? probing compositionality in large image models. In *Findings of the Association for Computational Linguistics: EACL 2024*, pages 1487–1500, 2024. 1, 2, 4

[56] Boyi Li, Kilian Q Weinberger, Serge Belongie, Vladlen Koltun, and Rene Ranftl. Language-driven semantic segmentation. In *International Conference on Learning Representations*, 2022. 1

[57] Junnan Li, Dongxu Li, Caiming Xiong, and Steven Hoi. Blip: Bootstrapping language-image pre-training for unified vision-language understanding and generation. In *International conference on machine learning*, pages 12888–12900. PMLR, 2022. 1

[58] Lin Li, Jun Xiao, Guikun Chen, Jian Shao, Yueting Zhuang, and Long Chen. Zero-shot visual relation detection via composite visual cues from large language models. *Advances in Neural Information Processing Systems*, 36, 2024. 6

[59] Rongjie Li, Songyang Zhang, Bo Wan, and Xuming He. Bipartite graph network with adaptive message passing for unbiased scene graph generation. In *Proceedings of the IEEE/CVF conference on computer vision and pattern recognition*, pages 11109–11119, 2021. 2

[60] Yangguang Li, Feng Liang, Lichen Zhao, Yufeng Cui, Wanli Ouyang, Jing Shao, Fengwei Yu, and Junjie Yan. Supervision exists everywhere: A data efficient contrastive language-image pre-training paradigm. In *International Conference on Learning Representations*, 2021. 1, 2

[61] Yanghao Li, Haoqi Fan, Ronghang Hu, Christoph Feichtenhofer, and Kaiming He. Scaling language-image pre-training via masking. In *Proceedings of the IEEE/CVF Conference on Computer Vision and Pattern Recognition*, pages 23390–23400, 2023. 2

[62] Chen Liang-Chieh, George Papandreou, Iasonas Kokkinos, Kevin Murphy, and Alan Yuille. Semantic image segmentation with deep convolutional nets and fully connected crfs. In *International conference on learning representations*, 2015. 2

[63] Haokun Lin, Haoli Bai, Zhili Liu, Lu Hou, Muyi Sun, Linqi Song, Ying Wei, and Zhenan Sun. Mope-clip: Structured pruning for efficient vision-language models with module-wise pruning error metric. In *Proceedings of the IEEE/CVF Conference on Computer Vision and Pattern Recognition*, pages 27370–27380, 2024. 1, 2

[64] Kevin Lin, Zhengyuan Yang, Linjie Li, Jianfeng Wang, and Lijuan Wang. Designbench: Exploring and benchmarking dall-e 3 for imagining visual design. *arXiv preprint arXiv:2310.15144*, 2023. 2, 5

[65] Tsung-Yi Lin, Michael Maire, Serge Belongie, James Hays, Pietro Perona, Deva Ramanan, Piotr Dollár, and

C Lawrence Zitnick. Microsoft coco: Common objects in context. In *Computer Vision–ECCV 2014: 13th European Conference, Zurich, Switzerland, September 6-12, 2014, Proceedings, Part V 13*, pages 740–755. Springer, 2014. 2, 3

[66] Tsung-Yi Lin, Piotr Dollár, Ross Girshick, Kaiming He, Bharath Hariharan, and Serge Belongie. Feature pyramid networks for object detection. In *Proceedings of the IEEE conference on computer vision and pattern recognition*, pages 2117–2125, 2017. 2

[67] Feng Liu, Minchul Kim, Zhiyuan Ren, and Xiaoming Liu. Distilling clip with dual guidance for learning discriminative human body shape representation. In *Proceedings of the IEEE/CVF Conference on Computer Vision and Pattern Recognition*, pages 256–266, 2024. 2

[68] Zhuang Liu, Hanzi Mao, Chao-Yuan Wu, Christoph Feichtenhofer, Trevor Darrell, and Saining Xie. A convnet for the 2020s. In *Proceedings of the IEEE/CVF conference on computer vision and pattern recognition*, pages 11976–11986, 2022. 8

[69] Yanxin Long, Youpeng Wen, Jianhua Han, Hang Xu, Pengzhen Ren, Wei Zhang, Shen Zhao, and Xiaodan Liang. Capdet: Unifying dense captioning and open-world detection pretraining. In *Proceedings of the IEEE/CVF Conference on Computer Vision and Pattern Recognition*, pages 15233–15243, 2023. 1

[70] Cewu Lu, Ranjay Krishna, Michael Bernstein, and Li Fei-Fei. Visual relationship detection with language priors. In *Computer Vision–ECCV 2016: 14th European Conference, Amsterdam, The Netherlands, October 11–14, 2016, Proceedings, Part I 14*, pages 852–869. Springer, 2016. 2

[71] Huaishao Luo, Lei Ji, Ming Zhong, Yang Chen, Wen Lei, Nan Duan, and Tianrui Li. Clip4clip: An empirical study of clip for end to end video clip retrieval and captioning. *Neurocomputing*, 508:293–304, 2022. 2

[72] Bolei Ma. Evaluating lexical aspect with large language models. In *Proceedings of the Workshop on Cognitive Modeling and Computational Linguistics*, pages 123–131, 2024. 4

[73] Dhruv Mahajan, Sundararajan Sellamanickam, and Vinod Nair. A joint learning framework for attribute models and object descriptions. In *2011 International Conference on Computer Vision*, pages 1227–1234. IEEE, 2011. 2

[74] Ségolène Martin, Yunshi Huang, Fereshteh Shakeri, Jean-Christophe Pesquet, and Ismail Ben Ayed. Transductive zero-shot and few-shot clip. In *Proceedings of the IEEE/CVF Conference on Computer Vision and Pattern Recognition*, pages 28816–28826, 2024. 2

[75] Yu Meng, Chenyan Xiong, Payal Bajaj, Paul Bennett, Jiawei Han, Xia Song, et al. Coco-lm: Correcting and contrasting text sequences for language model pretraining. *Advances in Neural Information Processing Systems*, 34: 23102–23114, 2021. 2

[76] Eleni Metheniti, Tim Van De Cruys, and Nabil Hathout. About time: Do transformers learn temporal verbal aspect? In *Proceedings of the Workshop on Cognitive Modeling and Computational Linguistics*, pages 88–101, Dublin, Ireland, 2022. Association for Computational Linguistics. 4

[77] Maitreya Patel, Abhiram Kusumba, Sheng Cheng, Changhoon Kim, Tejas Gokhale, Chitta Baral, and Yezhou Yang. Tripletclip: Improving compositional reasoning of clip via synthetic vision-language negatives. *Advances in neural information processing systems*, 2024. 2

[78] Genevieve Patterson and James Hays. Coco attributes: Attributes for people, animals, and objects. In *Computer Vision–ECCV 2016: 14th European Conference, Amsterdam, The Netherlands, October 11-14, 2016, Proceedings, Part VI 14*, pages 85–100. Springer, 2016. 2

[79] Sachin Pawar, Girish K Palshikar, and Pushpak Bhattacharyya. Relation extraction: A survey. *arXiv preprint arXiv:1712.05191*, 2017. 2

[80] Julia Peyre, Josef Sivic, Ivan Laptev, and Cordelia Schmid. Weakly-supervised learning of visual relations. In *Proceedings of the ieee international conference on computer vision*, pages 5179–5188, 2017. 2

[81] Bryan A Plummer, Arun Mallya, Christopher M Cervantes, Julia Hockenmaier, and Svetlana Lazebnik. Phrase localization and visual relationship detection with comprehensive image-language cues. In *Proceedings of the IEEE international conference on computer vision*, pages 1928–1937, 2017. 2

[82] Siyuan Qiao, Liang-Chieh Chen, and Alan Yuille. Detectors: Detecting objects with recursive feature pyramid and switchable atrous convolution. In *Proceedings of the IEEE/CVF conference on computer vision and pattern recognition*, pages 10213–10224, 2021. 2

[83] Alec Radford, Jong Wook Kim, Chris Hallacy, Aditya Ramesh, Gabriel Goh, Sandhini Agarwal, Girish Sastry, Amanda Askell, Pamela Mishkin, Jack Clark, et al. Learning transferable visual models from natural language supervision. In *International conference on machine learning*, pages 8748–8763. PMLR, 2021. 1, 2, 8

[84] Shuhuai Ren, Junyang Lin, Guangxiang Zhao, Rui Men, An Yang, Jingren Zhou, Xu Sun, and Hongxia Yang. Learning relation alignment for calibrated cross-modal retrieval. In *Proceedings of the 59th Annual Meeting of the Association for Computational Linguistics and the 11th International Joint Conference on Natural Language Processing (Volume 1: Long Papers)*, pages 514–524, 2021. 2

[85] Joshua Robinson, Ching-Yao Chuang, Suvrit Sra, and Stefanie Jegelka. Contrastive learning with hard negative samples. In *International Conference on Learning Representations (ICLR)*, 2021. 2

[86] MA Sadeghi and A Farhadi. Recognition using visual phrases. In *Proceedings of the 2011 IEEE Conference on Computer Vision and Pattern Recognition*, pages 1745–1752, 2011. 2

[87] Ugur Sahin, Hang Li, Qadeer Khan, Daniel Cremers, and Volker Tresp. Enhancing multimodal compositional reasoning of visual language models with generative negative mining. In *Proceedings of the IEEE/CVF Winter Conference on Applications of Computer Vision*, pages 5563–5573, 2024. 2

[88] Florian Schroff, Dmitry Kalenichenko, and James Philbin. Facenet: A unified embedding for face recognition and

clustering. In *Proceedings of the IEEE conference on computer vision and pattern recognition*, pages 815–823, 2015. 2

[89] Sebastian Schuster, Ranjay Krishna, Angel Chang, Li Fei-Fei, and Christopher D Manning. Generating semantically precise scene graphs from textual descriptions for improved image retrieval. In *Proceedings of the fourth workshop on vision and language*, pages 70–80, 2015. 3

[90] Shuai Shao, Zeming Li, Tianyuan Zhang, Chao Peng, Gang Yu, Xiangyu Zhang, Jing Li, and Jian Sun. Objects365: A large-scale, high-quality dataset for object detection. In *Proceedings of the IEEE/CVF international conference on computer vision*, pages 8430–8439, 2019. 2

[91] Viktoriia Sharmanska, Novi Quadrianto, and Christoph H Lampert. Augmented attribute representations. In *Computer Vision–ECCV 2012: 12th European Conference on Computer Vision, Florence, Italy, October 7-13, 2012, Proceedings, Part V 12*, pages 242–255. Springer, 2012. 2

[92] Sheng Shen, Liunian Harold Li, Hao Tan, Mohit Bansal, Anna Rohrbach, Kai-Wei Chang, Zhewei Yao, and Kurt Keutzer. How much can clip benefit vision-and-language tasks? In *International Conference on Learning Representations*, 2021. 2

[93] Amanpreet Singh, Ronghang Hu, Vedanuj Goswami, Guillaume Couairon, Wojciech Galuba, Marcus Rohrbach, and Douwe Kiela. Flava: A foundational language and vision alignment model. In *Proceedings of the IEEE/CVF Conference on Computer Vision and Pattern Recognition*, pages 15638–15650, 2022. 1, 2

[94] Jaisidh Singh, Ishaan Shrivastava, Mayank Vatsa, Richa Singh, and Aparna Bharati. Learn" no" to say" yes" better: Improving vision-language models via negations. *arXiv preprint arXiv:2403.20312*, 2024. 8

[95] Jasper Snoek, Hugo Larochelle, and Ryan P Adams. Practical bayesian optimization of machine learning algorithms. *Advances in neural information processing systems*, 25, 2012. 6

[96] Richard Socher, Brody Huval, Christopher D Manning, and Andrew Y Ng. Semantic compositionality through recursive matrix-vector spaces. In *Proceedings of the 2012 joint conference on empirical methods in natural language processing and computational natural language learning*, pages 1201–1211, 2012. 2

[97] Richard Socher, Milind Ganjoo, Christopher D Manning, and Andrew Ng. Zero-shot learning through cross-modal transfer. *Advances in neural information processing systems*, 26, 2013. 2

[98] Zeyi Sun, Ye Fang, Tong Wu, Pan Zhang, Yuhang Zang, Shu Kong, Yuanjun Xiong, Dahua Lin, and Jiaqi Wang. Alpha-clip: A clip model focusing on wherever you want. In *Proceedings of the IEEE/CVF Conference on Computer Vision and Pattern Recognition*, pages 13019–13029, 2024. 1, 2

[99] Ming Tao, Bing-Kun Bao, Hao Tang, and Changsheng Xu. Galip: Generative adversarial clips for text-to-image synthesis. In *Proceedings of the IEEE/CVF Conference on Computer Vision and Pattern Recognition*, pages 14214–14223, 2023. 1

[100] Shengbang Tong, Zhuang Liu, Yuexiang Zhai, Yi Ma, Yann LeCun, and Saining Xie. Eyes wide shut? exploring the visual shortcomings of multimodal llms. In *Proceedings of the IEEE/CVF Conference on Computer Vision and Pattern Recognition*, pages 9568–9578, 2024. 2, 4

[101] Yuli Vasiliev. *Natural language processing with Python and spaCy: A practical introduction*. No Starch Press, 2020. 3, 5

[102] Andreas Veit, Tomas Matera, Lukas Neumann, Jiri Matas, and Serge Belongie. Coco-text: Dataset and benchmark for text detection and recognition in natural images. *arXiv preprint arXiv:1601.07140*, 2016. 2

[103] Gang Wang and David Forsyth. Joint learning of visual attributes, object classes and visual saliency. In *2009 IEEE 12th International Conference on Computer Vision*, pages 537–544. IEEE, 2009. 2

[104] Huiyu Wang, Yukun Zhu, Hartwig Adam, Alan Yuille, and Liang-Chieh Chen. Max-deeplab: End-to-end panoptic segmentation with mask transformers. In *Proceedings of the IEEE/CVF conference on computer vision and pattern recognition*, pages 5463–5474, 2021. 2

[105] Yang Wang and Greg Mori. A discriminative latent model of object classes and attributes. In *Computer Vision–ECCV 2010: 11th European Conference on Computer Vision, Heraklion, Crete, Greece, September 5-11, 2010, Proceedings, Part V 11*, pages 155–168. Springer, 2010. 2

[106] Zhenyu Wang, Yali Li, Xi Chen, Ser-Nam Lim, Antonio Torralba, Hengshuang Zhao, and Shengjin Wang. Detecting everything in the open world: Towards universal object detection. In *Proceedings of the IEEE/CVF Conference on Computer Vision and Pattern Recognition*, pages 11433–11443, 2023. 1

[107] Longhui Wei, Lingxi Xie, Wengang Zhou, Houqiang Li, and Qi Tian. Mvp: Multimodality-guided visual pre-training. In *European conference on computer vision*, pages 337–353. Springer, 2022. 2

[108] Yongqin Xian, Bernt Schiele, and Zeynep Akata. Zero-shot learning-the good, the bad and the ugly. In *Proceedings of the IEEE conference on computer vision and pattern recognition*, pages 4582–4591, 2017. 2

[109] Danfei Xu, Yuke Zhu, Christopher B Choy, and Li Fei-Fei. Scene graph generation by iterative message passing. In *Proceedings of the IEEE conference on computer vision and pattern recognition*, pages 5410–5419, 2017. 2

[110] Jiarui Xu, Shalini De Mello, Sifei Liu, Wonmin Byeon, Thomas Breuel, Jan Kautz, and Xiaolong Wang. Groupvit: Semantic segmentation emerges from text supervision. In *Proceedings of the IEEE/CVF Conference on Computer Vision and Pattern Recognition*, pages 18134–18144, 2022. 1

[111] Chao Yi, Lu Ren, De-Chuan Zhan, and Han-Jia Ye. Leveraging cross-modal neighbor representation for improved clip classification. In *Proceedings of the IEEE/CVF Conference on Computer Vision and Pattern Recognition*, pages 27402–27411, 2024. 1, 2

[112] Jiahui Yu, Zirui Wang, Vijay Vasudevan, Legg Yeung, Mojtaba Seyedhosseini, and Yonghui Wu. Coca: Contrastive

captioners are image-text foundation models. *Transactions on Machine Learning Research*, 2022. 8

[113] Licheng Yu, Patrick Poirson, Shan Yang, Alexander C Berg, and Tamara L Berg. Modeling context in referring expressions. In *Computer Vision–ECCV 2016: 14th European Conference, Amsterdam, The Netherlands, October 11-14, 2016, Proceedings, Part II 14*, pages 69–85. Springer, 2016. 2

[114] Qihang Yu, Huiyu Wang, Dahun Kim, Siyuan Qiao, Maxwell Collins, Yukun Zhu, Hartwig Adam, Alan Yuille, and Liang-Chieh Chen. Cmt-deeplab: Clustering mask transformers for panoptic segmentation. In *Proceedings of the IEEE/CVF conference on computer vision and pattern recognition*, pages 2560–2570, 2022. 2

[115] Mert Yuksekgonul, Federico Bianchi, Pratyusha Kalluri, Dan Jurafsky, and James Zou. When and why vision-language models behave like bags-of-words, and what to do about it? *arXiv preprint arXiv:2210.01936*, 2022. 1, 2, 4, 7, 8

[116] Xiaohua Zhai, Basil Mustafa, Alexander Kolesnikov, and Lucas Beyer. Sigmoid loss for language image pre-training. In *Proceedings of the IEEE/CVF International Conference on Computer Vision*, pages 11975–11986, 2023. 6, 8

[117] Yibing Zhan, Jun Yu, Ting Yu, and Dacheng Tao. On exploring undetermined relationships for visual relationship detection. In *Proceedings of the IEEE/CVF Conference on Computer Vision and Pattern Recognition*, pages 5128–5137, 2019. 2

[118] Ji Zhang, Yannis Kalantidis, Marcus Rohrbach, Manohar Paluri, Ahmed Elgammal, and Mohamed Elhoseiny. Large-scale visual relationship understanding. In *Proceedings of the AAAI conference on artificial intelligence*, pages 9185–9194, 2019. 2

[119] Jingyi Zhang, Jiaxing Huang, Sheng Jin, and Shijian Lu. Vision-language models for vision tasks: A survey. *IEEE Transactions on Pattern Analysis and Machine Intelligence*, 2024. 2

[120] Le Zhang, Rabiul Awal, and Aishwarya Agrawal. Contrasting intra-modal and ranking cross-modal hard negatives to enhance visio-linguistic compositional understanding. In *Proceedings of the IEEE/CVF Conference on Computer Vision and Pattern Recognition*, pages 13774–13784, 2024. 2, 6, 8

[121] Richard Zhang. Making convolutional networks shift-invariant again. In *International conference on machine learning*, pages 7324–7334. PMLR, 2019. 4

[122] Renrui Zhang, Ziyu Guo, Wei Zhang, Kunchang Li, Xupeng Miao, Bin Cui, Yu Qiao, Peng Gao, and Hongsheng Li. Pointclip: Point cloud understanding by clip. In *Proceedings of the IEEE/CVF conference on computer vision and pattern recognition*, pages 8552–8562, 2022. 2

[123] Yi Zhang, Meng-Hao Guo, Miao Wang, and Shi-Min Hu. Exploring regional clues in clip for zero-shot semantic segmentation. In *Proceedings of the IEEE/CVF Conference on Computer Vision and Pattern Recognition*, pages 3270–3280, 2024. 2

[124] Long Zhao, Liangzhe Yuan, Boqing Gong, Yin Cui, Florian Schroff, Ming-Hsuan Yang, Hartwig Adam, and Ting Liu. Unified visual relationship detection with vision and language models. In *Proceedings of the IEEE/CVF International Conference on Computer Vision*, pages 6962–6973, 2023. 2, 6

[125] Tiancheng Zhao, Tianqi Zhang, Mingwei Zhu, Haozhan Shen, Kyusong Lee, Xiaopeng Lu, and Jianwei Yin. Vl-checklist: Evaluating pre-trained vision-language models with objects, attributes and relations. *arXiv preprint arXiv:2207.00221*, 2022. 2, 4

[126] Xubin Zhong, Xian Qu, Changxing Ding, and Dacheng Tao. Glance and gaze: Inferring action-aware points for one-stage human-object interaction detection. In *Proceedings of the IEEE/CVF Conference on Computer Vision and Pattern Recognition*, pages 13234–13243, 2021. 2

[127] Zexuan Zhong and Danqi Chen. A frustratingly easy approach for entity and relation extraction. In *Proceedings of the 2021 Conference of the North American Chapter of the Association for Computational Linguistics: Human Language Technologies*, pages 50–61, 2021. 2

[128] GuoDong Zhou, Jian Su, Jie Zhang, and Min Zhang. Exploring various knowledge in relation extraction. In *Proceedings of the 43rd annual meeting of the association for computational linguistics (acl'05)*, pages 427–434, 2005. 2

[129] Guodong Zhou, Min Zhang, DongHong Ji, and Qiaoming Zhu. Tree kernel-based relation extraction with context-sensitive structured parse tree information. In *Proceedings of the 2007 Joint Conference on Empirical Methods in Natural Language Processing and Computational Natural Language Learning (EMNLP-CoNLL)*, pages 728–736, 2007. 2

[130] Yufan Zhou, Ruiyi Zhang, Changyou Chen, Chunyuan Li, Chris Tensmeyer, Tong Yu, Jiuxiang Gu, Jinhui Xu, and Tong Sun. Lafite: Towards language-free training for text-to-image generation. In *Proceedings of the IEEE/CVF Conference on Computer Vision and Pattern Recognition (CVPR)*, 2022. 1

[131] Yi Zhu, Zhaoqing Zhu, Bingqian Lin, Xiaodan Liang, Feng Zhao, and Jianzhuang Liu. Relclip: Adapting language-image pretraining for visual relationship detection via relational contrastive learning. In *Proceedings of the 2022 Conference on Empirical Methods in Natural Language Processing*, pages 4800–4810, 2022. 2